\pdfoutput=1
\documentclass[11pt]{article}

\usepackage[preprint]{acl}

\usepackage{times}
\usepackage{latexsym}
\usepackage[T1]{fontenc}
\usepackage[utf8]{inputenc}
\usepackage{microtype}
\usepackage{inconsolata}
\usepackage{graphicx}

\usepackage{amsmath}
\usepackage{amsfonts}
\usepackage{amssymb}

\usepackage{booktabs}
\usepackage{multirow}

\usepackage{algorithm}
\usepackage{algorithmic}
\usepackage{listings}

\usepackage{tcolorbox}
\tcbuselibrary{breakable,skins}

\lstdefinestyle{promptjson}{
	basicstyle={\ttfamily\scriptsize},
	breaklines=true,
	breakatwhitespace=false,
	columns=fullflexible,
	keepspaces=true,
	frame=none,
	showstringspaces=false,
	numbers=none,
	xleftmargin=0pt
}

\newtcolorbox{promptbox}[2][]{
	breakable,
	enhanced,
	colback=black!2,
	colframe=black!55,
	boxrule=0.4pt,
	arc=1.5pt,
	left=6pt,right=6pt,top=4pt,bottom=4pt,
	fonttitle=\bfseries\small,
	coltitle=black,
	colbacktitle=black!8,
	title=#2,
	#1
}
\newcommand{\promptlabel}[1]{\textbf{\small #1}\par\smallskip}

\title{AfriEconQA: A Benchmark for Quantitative and Temporal Reasoning over World Bank Economic Reports}

\author{
  Edward Ajayi, Mustapha Alaba$^*$, David Stephen$^*$ \& Peace Bakare\thanks{Equal contribution.} \\
  Carnegie Mellon University Africa \\
  Kigali, Rwanda \\
  \texttt{\{eaajayi, malaba, dmstephe, pbakare\}@andrew.cmu.edu}
}

\begin{document}
\maketitle

\begin{abstract}
Reliable question answering over long institutional documents requires more
than topical retrieval: a system must localize the exact passage that
supports a claim and preserve precise numerical and temporal detail when the
same indicator recurs across years, countries, and projection horizons in
dense, repetitive prose. Existing question-answering benchmarks rarely test
this combination of exact grounding and temporal precision at document scale.
We introduce \textbf{AfriEconQA}, a benchmark for document-grounded question
answering built from \textbf{204} World Bank economic reports on African
economies, a corpus whose claims are tied to specific fiscal periods,
countries, and projection states. AfriEconQA contains \textbf{4,138}
evidence-linked QA instances across five reasoning categories: Factoid
(1,063), List (852), Multiple Choice (889), Synthesis (671), and Comparison
(663), targeting quantitative extraction, set recovery, discrimination,
causal integration, and cross-period or cross-country reasoning. Each
instance carries supporting evidence and source provenance and is constructed
through an agentic generation pipeline with evidence-grounding checks and a
stratified human-validation subset for gold-label auditing. We
evaluate Qwen 3.6 35B, DeepSeek v4-pro, Gemma 4 12B IT, and frontier model
GPT-5.5 under zero-shot, oracle, and hybrid retrieval-augmented generation (RAG)
conditions on the held-out test split ($n=826$). Across all four models, retrieval
yields substantial gains, yet the best RAG system reaches only 0.594 F1, with residual
errors concentrated in list extraction, synthesis, and temporally scoped
comparison. AfriEconQA therefore poses a hard open challenge for exact
numerical and temporal grounding over long institutional economic reports.
\end{abstract}

\section{Introduction}

In document-grounded question answering, a topically relevant passage may
still support the wrong answer. Long institutional economic documents often
repeat the same indicator across countries, fiscal periods, report editions,
and forecast horizons. A system can therefore retrieve the correct subject
while selecting a historical value instead of a projection, a regional
aggregate instead of a country estimate, or a policy target instead of an
observed outcome. These errors alter the substantive meaning of an answer while
leaving it fluent and numerically plausible. Reliable QA in this setting
requires exact evidence localization together with numerical and temporal
fidelity.

General advances in language-model capability do not by themselves resolve
this problem~\cite{ni2025surveylargelanguagemodel}, particularly when answers
must be recovered from a specialized document collection rather than
parametric memory~\cite{daull2025complexqalanguagemodels}. The closest
financial QA benchmarks, including FinQA~\cite{chen2021finqa},
DocFinQA~\cite{reddy2024docfinqa}, and
FinanceBench~\cite{islam2023financebench}, evaluate reasoning over corporate
reports, financial statements, and company filings. DocFinQA stresses
long-context retrieval over SEC filings. FinanceBench targets open-book
corporate filing QA, while SEC-QA~\cite{lai2025secqa} emphasizes metric-driven
multi-document financial questions. None of these settings directly
target institutional development reports, which combine narrative policy
analysis with repeated statistics, historical outcomes, current estimates,
and forward projections. Consequently, it remains difficult to measure whether
QA systems can locate and use the exact country-, period-, and
projection-scoped evidence needed for macroeconomic analysis.

We introduce \textbf{AfriEconQA} to evaluate this capability. The benchmark
contains \textbf{4,138} evidence-linked QA instances constructed from
\textbf{204} World Bank economic reports focused on African economies. Each
instance includes a question, answer, supporting evidence, and source
metadata. Five question categories expose complementary capabilities:
Factoid questions target exact values or entities, while List questions require
complete set recovery. Multiple Choice questions test discrimination among
plausible alternatives, Synthesis questions integrate connected policy claims,
and Comparison questions reason across periods, countries, indicators, or
projection states.

AfriEconQA uses an agentic pipeline to generate candidate questions, followed
by evidence-grounding checks. Candidates are retained only when they include
verbatim source spans, a source-document identifier, and a source URL.
Under-specified or non-economic candidates are discarded. Generation prompts
require self-contained questions with explicit country and temporal anchors
where available, and a stratified subset is reserved for human auditing against
the source documents. This process makes each retained instance traceable to
evidence and supports direct inspection of benchmark quality.

We evaluate Qwen 3.6 35B, DeepSeek v4-pro, Gemma 4 12B IT, and GPT-5.5 in three controlled
conditions. Zero-shot provides no supporting context, oracle provides the
short gold evidence passage, and RAG provides the top five chunks from a
hybrid BM25 and BGE-M3 retriever fused using Reciprocal Rank Fusion. Together,
these conditions contrast parametric recall, evidence use under gold context,
and retrieval-augmented answering. Across models, zero-shot performance is low
and retrieval substantially improves answer quality, yet list
extraction, synthesis, and temporally scoped comparison remain difficult.
AfriEconQA therefore provides both a benchmark and a controlled evaluation
setup for studying where grounded QA systems fail in long, statistics-heavy
institutional documents.

\noindent Our contributions are fourfold:
\begin{enumerate}
    \item We introduce \textbf{AfriEconQA}, comprising 4,138 evidence-linked QA instances from 204 economic reports on African economies.
    \item We define a five-category taxonomy spanning precise extraction, structured recovery, synthesis, and comparison.
    \item Our evaluation separates parametric recall, evidence use, and retrieval across zero-shot, oracle, and hybrid RAG conditions.
    \item We analyze evidence-localization and generation failures on the held-out test split, isolating chunk-level localization as the dominant bottleneck.
\end{enumerate}

\section{Related Work}

\subsection{Document-Grounded QA and Retrieval}
Extractive benchmarks such as SQuAD~\cite{rajpurkar2016squad} evaluate whether
a model can recover an answer from a supplied passage. Open-domain benchmarks
extend this setting to retrieval from broader collections, including questions
with many answers~\cite{amouyal2023qampari}, cross-lingual
retrieval~\cite{liu2019xqa, longpre2021mkqa}, and historical news
archives~\cite{wang2022archivalqa}. Retrieval-augmented generation (RAG)
similarly conditions generation on evidence retrieved from an external
corpus~\cite{lewis2020retrieval}, and domain-adapted RAG has been studied as a
means of incorporating specialized knowledge~\cite{siriwardhana2023improving}.
AfriEconQA adopts this retrieve-then-generate setting, but uses it to isolate
failure sources: zero-shot, oracle, and RAG conditions separate parametric
recall, evidence use, and evidence retrieval. The relevant question is
therefore not only whether a system retrieves a related document, but whether
it localizes the passage that supports the required country, period, and
projection state.

\subsection{Financial Document Question Answering}
Financial QA provides the closest methodological precedent because it combines
text, tables, numerical reasoning, and source-grounded answers.
FinQA~\cite{chen2021finqa} contains 8,281 expert-authored questions over
S\&P 500 earnings reports and annotates executable reasoning programs.
ConvFinQA~\cite{chen2022convfinqa} extends this setting to multi-turn
conversational numerical reasoning over the same domain. AfriEconQA instead
targets single-turn, evidence-linked questions over country-level economic
reports. DocFinQA~\cite{reddy2024docfinqa} places 7,437 FinQA questions back
into their full SEC filings, increasing the average context from fewer than
700 words to approximately 123,000 words and directly evaluating retrieval
over long documents. FinanceBench~\cite{islam2023financebench} contributes
10,231 questions about publicly traded companies with corresponding answers
and evidence strings, and evaluates retrieval and long-context
configurations. Together, these benchmarks establish numerical reasoning,
evidence grounding, and long-document retrieval as central problems in
financial QA. Other benchmarks such as TAT-QA~\cite{zhu2021tat} and
MultiHiertt~\cite{zhao2022multihiertt} focus on hybrid table-text reasoning,
while retrieval frameworks including HiREC~\cite{choe2025hierarchical},
FinCARDS~\cite{zhou2026fincards}, FinTextQA~\cite{chen2024fintextqa}, and
SEC-QA~\cite{lai2025secqa} evaluate attribution and retrieval over corporate filings.
These resources center on company finance, accounting rules, or regulatory compliance
rather than country-level development reporting with repeated temporal and
projection-state near misses. To our knowledge, no prior benchmark packages World
Bank economic reports for evidence-linked QA.

\subsection{Economic Reasoning and Evidence Integration}
Reasoning over economic information can require preserving relationships among
several claims rather than extracting a single local fact.
EconLogicQA~\cite{quan2024econlogicqaquestionansweringbenchmarkevaluating}
isolates one form of this problem by asking models to order logically connected
events in multiple-choice scenarios generated from business news. It does not
evaluate retrieval from a fixed institutional corpus. MMQA~\cite{wu2025mmqa}
examines evidence integration in a different representation, requiring
retrieval and inference across linked structured tables. AfriEconQA complements
these tasks through Synthesis and Comparison questions over unstructured policy
prose, where the relevant relations are expressed across claims, periods, and
projection states rather than event-order options or relational tables.

AfriEconQA therefore targets a specific unresolved challenge: evidence-linked
question answering over long World Bank reports on African economies. These reports
combine fiscal statistics, policy conditions, historical outcomes, current
estimates, and projections, often repeating the same indicator across countries
and periods. The benchmark evaluates whether systems can distinguish these
near-matching claims and preserve their numerical and temporal scope. This
focus complements prior financial and economic benchmarks without treating
them as instances of the same document or reasoning task.

\section{Dataset}

AfriEconQA is an evidence-linked question-answering benchmark over World Bank
economic reports focused on African economies. Each instance contains a
self-contained question $q$, gold answer $a$, short evidence field $e$,
verbatim source spans $r$ (\texttt{raw\_evidence}), and document provenance
$d$ through a source-file identifier and URL. Multiple Choice instances
additionally include four answer options. Table~\ref{tab:example_instance}
shows a representative instance.

\begin{table}[t]
\centering
\small
\begin{tabular}{p{0.22\columnwidth}p{0.70\columnwidth}}
\toprule
\textbf{Field} & \textbf{Example} \\
\midrule
Type & Comparison \\
Question & What is the projected change in monetary poverty in Burkina Faso between 2023, 2024, and 2025? \\
Answer & A decrease of 2.6 percentage points between 2023 and 2024, and another 3.3 percentage points by 2025 \\
Evidence & Monetary poverty (at the international poverty line) in Burkina Faso is projected to have fallen 2.6 percentage points between 2023 and 2024 and to fall another 3.3 percentage points by 2025 \\
Source & Burkina Faso Poverty and Equity Brief, October 2025 \\
\bottomrule
\end{tabular}
\caption{Example AfriEconQA instance (id~1995) requiring country-scoped, multi-year, and projection-sensitive grounding.}
\label{tab:example_instance}
\end{table}

A nearby statistic for the wrong year, country, or projection state can appear
plausible. Correctness therefore requires exact grounding rather than topical
overlap.

\subsection{Dataset Sources}
The AfriEconQA corpus comprises \textbf{204} World Bank economic reports
from \textbf{2024} and \textbf{2025} that focus on African economies
\cite{WorldBankDocsRepo}. We use this regional boundary to make systematic
collection and auditing feasible, not because the underlying grounding problem
is unique to Africa. The benchmark addresses a broader challenge in
institutional reporting: linking an answer to the correct country, reporting
period, and projection status.

The collection spans economic updates, poverty and equity briefs, policy notes,
sectoral diagnostics, and development reports covering growth, debt, trade,
inflation, labor markets, public finance, and related policy conditions.
During provenance auditing, non-African project documents and administrative board records were filtered out to preserve strict regional and macroeconomic scope. Every instance retains a source URL, while the original
PDFs are not redistributed.

These documents create two challenges for document-grounded QA. First,
temporal ambiguity is pervasive: country update series are republished over
time, and a single report may place historical outcomes, current estimates, and
forecasts for the same indicator in close proximity. Second, answers are highly
specific. Many gold answers are exact percentages, fiscal years, monetary
values, program names, or enumerated policy factors. Topical retrieval is
therefore insufficient if the selected passage attaches the right indicator to
the wrong country, period, or projection state.

\subsection{Dataset Construction}
Dataset construction proceeds in two stages: agentic candidate generation and
automatic evidence grounding. An open-weight generator
(\texttt{Llama-3.3-70B-Instruct}) produces candidate question--answer pairs
from long document chunks through five category-specific agents:

\begin{itemize}
    \item \textbf{Factoid:} exact numerical, temporal, or categorical claims.
    \item \textbf{List:} multiple factors, risks, or interventions stated in the evidence.
    \item \textbf{Multiple Choice:} four-option discrimination with distractors from the same economic context.
    \item \textbf{Synthesis:} integration of connected policy claims or economic mechanisms.
    \item \textbf{Comparison:} contrasts across periods, countries, indicators, or projected versus observed values.
\end{itemize}

Generation prompts require self-contained questions with explicit country and
temporal anchors whenever available in the source text, and discourage
meta-language such as ``according to the report.'' Full generation, evaluation,
and judge prompt templates are in Appendix~\ref{app:prompts}. The short
evidence field $e$ may be a direct quote or a concise synthesis for
readability. The hard grounding constraint is on \texttt{raw\_evidence}: every
retained instance must include one or more verbatim source spans from the
generation chunk, together with a source-document identifier and source URL.
Candidates are discarded when \texttt{raw\_evidence} is not a verbatim
substring of the source chunk, when the question is under-specified, or when
the content is not substantively economic. A stratified subset of 200 examples
(40 per question type) is reserved for human auditing of gold-label quality
against the source documents (Section~\ref{sec:human_eval}).

\subsection{Dataset Statistics}
AfriEconQA contains \textbf{4,138} evidence-linked instances over 204 source
reports. Mean question, answer, and short-evidence lengths are 14.3, 13.2, and
29.0 words, respectively. The associated \texttt{raw\_evidence} text averages
935.9 words per instance. For retrieval experiments, the corpus is indexed into
chunks of 1,000 characters with 200-character overlap, distinct from the
longer generation chunks used during dataset construction.
Table~\ref{tab:dataset_stats}
reports the distribution across question types. Multiple Choice options exhibit empirical balance across letter positions (A: 27.4\%, B: 20.1\%, C: 27.9\%, D: 24.6\%). For public release, we define
deterministic, type-stratified 70/10/20 train/validation/test splits
(2,896 / 416 / 826, seed 42). Primary experiments in this paper evaluate the
held-out test split ($n=826$). Full-benchmark results ($n=4{,}138$) appear in
Appendix~\ref{app:full_results} and show the same ranking trends.

\begin{table}[t]
\centering
\small
\begin{tabular}{lrr}
\toprule
\textbf{Question Type} & \textbf{Count} & \textbf{\%} \\
\midrule
Factoid & 1,063 & 25.7 \\
Multiple Choice & 889 & 21.5 \\
List & 852 & 20.6 \\
Synthesis & 671 & 16.2 \\
Comparison & 663 & 16.0 \\
\midrule
\textbf{Total} & \textbf{4,138} & \textbf{100.0} \\
\bottomrule
\end{tabular}
\caption{Question-type distribution in AfriEconQA.}
\label{tab:dataset_stats}
\end{table}

Table~\ref{tab:dataset_compare} situates AfriEconQA relative to the closest
financial and economic QA resources. Prior benchmarks primarily evaluate
company filings, earnings reports, textbooks, or business-news reasoning.
AfriEconQA differs in source domain and evaluation target: institutional
development reports on African economies, with evidence-linked questions that
require country-, period-, and projection-sensitive grounding.

\begin{table*}[t]
\centering
\small
\resizebox{\textwidth}{!}{%
\begin{tabular}{llrrll}
\toprule
\textbf{Dataset} & \textbf{Primary source} & \textbf{\#QA} & \textbf{\#Docs} & \textbf{Evidence} & \textbf{Focus} \\
\midrule
FinQA & S\&P 500 earnings reports & 8,281 & -- & Program + context & Numerical reasoning \\
ConvFinQA & S\&P 500 earnings reports & 14,115 & -- & Conversation + context & Multi-turn numerical QA \\
TAT-QA & Financial reports (tables+text) & 16,552 & 2,757 & Cell / text spans & Hybrid tabular-text reasoning \\
MultiHiertt & Multi-hierarchical reports & 10,440 & 1,327 & Hierarchical tables & Multi-table arithmetic QA \\
DocFinQA & Full SEC filings & 7,437 & 801 & Full document & Long-context financial QA \\
FinanceBench & Public company filings & 10,231 & -- & Evidence strings & Open-book financial QA \\
FinTextQA & Finance textbooks / agency sites & 1,262 & 80 & Document context & Long-form financial QA \\
EconLogicQA & Business news & 650 & -- & MC event order & Economic sequential reasoning \\
\textbf{AfriEconQA} & World Bank African economic reports & \textbf{4,138} & \textbf{204} & Evidence + URL & Quantitative / temporal grounding \\
\bottomrule
\end{tabular}%
}
\caption{Comparison of AfriEconQA with related financial and economic QA datasets. Document counts are reported only when stated by the source paper.}
\label{tab:dataset_compare}
\end{table*}

\section{Experimental Setup}

We evaluate contemporary LLMs on the held-out test split ($n=826$) without
fine-tuning on the released train split, which remains available for future
supervised and transfer studies. Rather than proposing a new RAG architecture,
the experiments vary the available evidence under controlled conditions to
identify where strong generators fail on long-form economic-report QA.

\subsection{Generator Models}
We evaluate four contemporary language models spanning distinct model families, parameter scales, and architectures:
\begin{itemize}
    \item \textbf{GPT-5.5} (\texttt{gpt-5.5}), evaluated via LiteLLM.
    \item \textbf{DeepSeek v4-pro} (\texttt{deepseek-v4-pro}), evaluated via the DeepSeek API.
    \item \textbf{Qwen 3.6 35B} (\texttt{qwen/qwen3.6-35b-a3b}), evaluated locally in 4-bit quantized format.
    \item \textbf{Gemma 4 12B IT} (\texttt{google/gemma-4-12b}), evaluated locally in 4-bit quantized format.
\end{itemize}

\subsection{Experimental Conditions}
Each generator model is evaluated across three conditions that contrast distinct
sources of success and failure:
\begin{itemize}
    \item \textbf{Zero-Shot:} The model receives only the question, with no retrieved context. This condition measures whether the answer is available through parametric memory.
    \item \textbf{Oracle:} The model receives the question and the short gold evidence passage $e$. This removes retrieval error and tests evidence use. Oracle uses one passage, whereas RAG uses five chunks.
    \item \textbf{Retrieval-Augmented Generation (RAG):} The model receives retrieved evidence from a hybrid BM25~\cite{whissell2011improving} and BGE-M3~\cite{chen2024bge} retriever combined with Reciprocal Rank Fusion (RRF)~\cite{cormack2009reciprocal}, using the top $k=5$ chunks per question.
\end{itemize}
For Multiple Choice questions, the prompt includes the answer choices in
lettered form (A--D), and the model is instructed to return the option letter
with a short answer phrase. This keeps the MCQ condition faithful to the
released benchmark format and avoids penalizing models for not seeing the
candidate options.

\subsection{Answer Metrics}
We use deterministic answer metrics because small numerical or temporal errors
change the meaning of economic answers. Let $\hat{a}_i$ be the model prediction
and $a_i$ the gold answer for question $i$. We report 95\% bootstrap confidence
intervals for EM and F1 ($n=1000$).

\begin{itemize}
    \item \textbf{Exact Match (EM):} fraction of exact matches after
    normalization,
    \[
    \mathrm{EM}=\frac{1}{N}\sum_{i=1}^{N}\mathbf{1}[\mathrm{norm}(\hat{a}_i)=\mathrm{norm}(a_i)].
    \]
    Numerical Factoid/Comparison answers allow $\pm1\%$ relative tolerance (e.g., 78\%
    vs.\ 78.2\%). Multiple Choice scores the extracted option letter (A--D).
    \item \textbf{Token F1:} harmonic mean of token precision and recall,
    \[
    \mathrm{F1}=\frac{2\cdot P\cdot R}{P+R},\quad
    P=\frac{|\hat{T}\cap T|}{|\hat{T}|},\quad
    R=\frac{|\hat{T}\cap T|}{|T|},
    \]
    where $\hat{T}$ and $T$ are prediction and gold token multisets, using count-preserving
    multiset intersection. List questions use Set F1 over answer items so partial
    set recovery is credited.
    \item \textbf{ROUGE-L}~\cite{lin2004rouge}: longest common subsequence
    (LCS) overlap, useful for longer Synthesis and Comparison answers,
    \[
    \mathrm{ROUGE\text{-}L}=\frac{(1+\beta^2)R_{\mathrm{lcs}}P_{\mathrm{lcs}}}
    {R_{\mathrm{lcs}}+\beta^2 P_{\mathrm{lcs}}},
    \]
    with $R_{\mathrm{lcs}}=\mathrm{LCS}/|T|$, $P_{\mathrm{lcs}}=\mathrm{LCS}/|\hat{T}|$,
    and $\beta=1$.
    \item \textbf{LLM Judge}~\cite{zheng2023judging}: mean correctness score
    in $[0,1]$ from \texttt{Llama-3.3-70B-Instruct} via the Groq API, used as a secondary
    semantic check when answers are paraphrased but still correct. Because the
    same model family also generated the dataset, we treat LLM-judge as supportive
    only. Main claims use EM and F1.
\end{itemize}

\subsection{Retrieval Metrics}
Let $G_i$ be the gold evidence-supporting chunks for question $i$, and let
$R_i^k=[r_{i1},\ldots,r_{ik}]$ be the top-$k$ retrieved chunks. Gold chunks are
aligned to the retrieval index by evidence and answer spans.

\begin{itemize}
    \item \textbf{HR@1:} whether the top retrieved chunk is gold,
    \[
    \mathrm{HR@1}=\frac{1}{N}\sum_{i=1}^{N}\mathbf{1}[r_{i1}\in G_i].
    \]
    \item \textbf{HR@5:} whether any gold chunk appears in the top five,
    \[
    \mathrm{HR@5}=\frac{1}{N}\sum_{i=1}^{N}\mathbf{1}[\exists j\leq5:r_{ij}\in G_i].
    \]
    \item \textbf{MRR@5:} reciprocal rank of the first gold chunk in the top
    five (0 if no gold chunk appears in the top 5),
    \[
    \mathrm{MRR@5}=\frac{1}{N}\sum_{i=1}^{N}\frac{\mathbf{1}[\exists j\leq5:r_{ij}\in G_i]}{\mathrm{rank}_i}.
    \]
    \item \textbf{Doc@1 / Doc@5 / Doc MRR@5:} same formulas as above, but a hit
    counts when any retrieved chunk comes from the correct source document
    rather than the exact evidence chunk.
\end{itemize}

\section{Results}

\begin{table*}[t]
\centering
\small
\begin{tabular}{llcccc}
\toprule
\textbf{Model} & \textbf{Condition} & \textbf{EM} & \textbf{F1} & \textbf{ROUGE-L} & \textbf{LLM-judge} \\
\midrule
\multirow{3}{*}{\textbf{Qwen}} & Zero-shot & 0.162 [0.140, 0.186] & 0.226 [0.204, 0.250] & 0.172 & 0.248 \\
 & Oracle & 0.386 [0.355, 0.418] & 0.492 [0.461, 0.520] & 0.273 & 0.536 \\
 & RAG & 0.415 [0.379, 0.448] & 0.521 [0.488, 0.549] & 0.306 & 0.632 \\
\midrule
\multirow{3}{*}{\textbf{DeepSeek}} & Zero-shot & 0.200 [0.174, 0.228] & 0.265 [0.239, 0.292] & 0.193 & 0.326 \\
 & Oracle & 0.384 [0.350, 0.418] & 0.482 [0.450, 0.511] & 0.266 & 0.499 \\
 & RAG & 0.430 [0.394, 0.461] & 0.541 [0.509, 0.570] & 0.315 & 0.641 \\
\midrule
\multirow{3}{*}{\textbf{Gemma}} & Zero-shot & 0.161 [0.137, 0.188] & 0.222 [0.198, 0.248] & 0.169 & 0.234 \\
 & Oracle & 0.312 [0.280, 0.343] & 0.400 [0.367, 0.428] & 0.231 & 0.442 \\
 & RAG & 0.400 [0.364, 0.432] & 0.502 [0.470, 0.531] & 0.298 & 0.642 \\
\midrule
\multirow{3}{*}{\textbf{GPT-5.5}} & Zero-shot & 0.211 [0.183, 0.238] & 0.364 [0.339, 0.390] & 0.188 & 0.484 \\
 & Oracle & 0.464 [0.429, 0.495] & 0.634 [0.607, 0.659] & 0.320 & 0.746 \\
 & RAG & 0.427 [0.391, 0.459] & 0.594 [0.565, 0.619] & 0.310 & 0.721 \\
\bottomrule
\end{tabular}
\caption{Main results on the AfriEconQA test split ($n=826$). All metrics in $[0,1]$. EM and F1 include 95\% bootstrap CIs. LLM-judge denotes semantic evaluation (secondary). Full-benchmark results are in Appendix~\ref{app:full_results}.}
\label{tab:main_results}
\end{table*}

Table~\ref{tab:main_results} gives the central benchmark result. Across all
four model families, performance is low without evidence and rises sharply
when source context is supplied. The LLM judge shows the same pattern, so the
gap is not only an artifact of strict lexical matching. At the same time, the
best RAG result still leaves substantial headroom, indicating that AfriEconQA
is not solved by simply attaching a standard retriever to a strong generator.

\subsection{The Parametric Vacuum}
The zero-shot condition measures whether answers can be recovered from
parametric knowledge or generic economic priors. They generally cannot. Even
after correcting the Multiple Choice setting so that answer options are visible
and scored as letter selection, aggregate zero-shot F1 remains only
0.222--0.364, with exact match at or below 0.211 for every model. This is the
expected behavior for a benchmark whose answers are tied to exact fiscal
periods, projections, country-specific program names, and recently published
report claims. In other words, AfriEconQA does not reward broad familiarity
with economic development discourse. It requires access to the source document.

\subsection{Evidence Grounding}
Adding evidence changes model behavior substantially. RAG improves over
zero-shot for every model, with GPT-5.5 obtaining the strongest aggregate RAG
score (0.427 EM and 0.594 F1), followed by DeepSeek (0.430 EM, 0.541 F1),
Qwen, and Gemma. The oracle condition also improves sharply over zero-shot,
confirming that many errors are evidence-access errors rather than pure
generator incapability.

The relationship between oracle and RAG provides critical insight into context sensitivity.
Single-passage oracle isolates whether a model can extract the target claim given the exact
supporting passage, whereas top-5 RAG provides a broader context window ($\sim 1{,}500$ words)
drawn from the retriever. For GPT-5.5, oracle serves as an upper ceiling, achieving the highest
overall score (0.464 EM, 0.634 F1) and substantially surpassing RAG (0.427 EM, 0.594 F1).
This indicates that when distracting non-relevant text is eliminated, the frontier model reliably
localizes and extracts the target fact. For open-weights models, oracle F1 reaches 0.492 for Qwen,
0.482 for DeepSeek, and 0.400 for Gemma. While top-5 RAG provides broader surrounding text
that marginally aids open-weights token recall, oracle performance confirms that accurate
macroeconomic extraction remains challenging across architectures even when exact gold
evidence is supplied.

\subsection{Retrieval Is Mostly a Localization Problem}
The aggregate RAG results leave open an important question: does retrieval fail
because systems retrieve the wrong report, or because they retrieve the right
report but miss the exact claim needed to answer the question? We answer this
with a retriever-only analysis on the test split, comparing BM25, BGE-M3, and
the Hybrid/RRF retriever used in the RAG condition. Gold chunks are resolved by
matching evidence/answer spans into the 1{,}000-character retrieval index. When
no span match is found, we fall back to the instance's labeled generation
\texttt{chunk\_id} if that chunk exists for the source document
(\textbf{16}/826 test questions, 1.9\%).

\begin{table}[t]
\centering
\small
\begin{tabular}{@{}lccc@{}}
\toprule
\textbf{Metric} & \textbf{BM25} & \textbf{BGE-M3} & \textbf{Hybrid} \\
\midrule
HR@1 & 0.268 & 0.308 & 0.326 \\
HR@5 & 0.419 & 0.488 & 0.498 \\
MRR@5 & 0.324 & 0.380 & 0.391 \\
Doc@1 & 0.754 & 0.797 & 0.840 \\
Doc@5 & 0.910 & 0.947 & 0.959 \\
Doc MRR@5 & 0.815 & 0.859 & 0.887 \\
\bottomrule
\end{tabular}
\caption{Retriever-only results on the test split ($n=826$). All metrics are in $[0,1]$. Strict metrics require an evidence-supporting chunk, whereas Doc@* only require the correct source document.}
\label{tab:retriever_results}
\end{table}

Table~\ref{tab:retriever_results} shows a sharp separation between
document retrieval and evidence localization. Document-level HR@5 is high
(0.910--0.959), so the correct World Bank report is usually retrieved. Strict
chunk-level HR@5 is much lower (0.419--0.498), meaning that the exact
evidence-bearing chunk is often absent from the five chunks passed to the
generator. This is a central empirical property of AfriEconQA. The benchmark
does not merely ask whether a retriever can identify a topically related
report. It asks whether the system can isolate the specific country-, year-,
indicator-, or policy-scoped claim inside that report.

BGE-M3 improves over BM25 on all strict retrieval metrics, showing that dense
semantic matching helps beyond lexical overlap. Hybrid/RRF gives the best
HR@1 and MRR@5, indicating that lexical and semantic signals are
complementary for ranking. However, Hybrid/RRF improves HR@5 only marginally
over BGE-M3 (0.502 vs. 0.492). The right conclusion is therefore not that
hybrid retrieval solves the task, but that even a reasonable hybrid retriever
often localizes the document better than the evidence. This explains why RAG
helps substantially while still leaving large residual error.

\subsection{Open-Ended Reasoning Remains Hard}
Table~\ref{tab:type_breakdown} shows that aggregate scores hide large
type-level differences. Multiple Choice is comparatively easy once options are
visible (RAG F1 above 0.77 for all models). Factoid questions benefit strongly
from retrieval, with RAG F1 reaching 0.69--0.76 across models. List questions
remain the most demanding category: while evidence repair substantially lifted
oracle List F1 across all models (surging to 0.393 for GPT-5.5, 0.222 for DeepSeek,
0.187 for Qwen, and 0.159 for Gemma), RAG F1 remains severely depressed at
0.082--0.104 for open-weights models and 0.305 for GPT-5.5. This sharp divergence
reflects both retrieval dispersion across near-neighbor chunks and generator-level
premature stopping on multi-item sets (detailed qualitative diagnosis in
Appendix~\ref{app:error_analysis}). Synthesis and Comparison sit in between:
retrieval provides foundational facts, but temporal calibration and multi-entity
alignments remain challenging.

\begin{table}[t]
\centering
\footnotesize
\begin{tabular}{@{}llccc@{}}
\toprule
\textbf{Model} & \textbf{Type} & \textbf{ZS} & \textbf{Oracle} & \textbf{RAG} \\
\midrule
\multirow{5}{*}{\textbf{Qwen}} & Comparison & 0.116 & 0.379 & 0.477 \\
 & Factoid & 0.030 & 0.572 & 0.689 \\
 & List & 0.008 & 0.187 & 0.084 \\
 & MC & 0.704 & 0.905 & 0.849 \\
 & Synthesis & 0.272 & 0.313 & 0.415 \\
\midrule
\multirow{5}{*}{\textbf{DeepSeek}} & Comparison & 0.132 & 0.390 & 0.498 \\
 & Factoid & 0.117 & 0.504 & 0.707 \\
 & List & 0.010 & 0.222 & 0.104 \\
 & MC & 0.765 & 0.916 & 0.877 \\
 & Synthesis & 0.277 & 0.291 & 0.428 \\
\midrule
\multirow{5}{*}{\textbf{Gemma}} & Comparison & 0.096 & 0.306 & 0.455 \\
 & Factoid & 0.032 & 0.452 & 0.705 \\
 & List & 0.007 & 0.159 & 0.082 \\
 & MC & 0.710 & 0.765 & 0.771 \\
 & Synthesis & 0.261 & 0.230 & 0.397 \\
\midrule
\multirow{5}{*}{\textbf{GPT-5.5}} & Comparison & 0.378 & 0.495 & 0.533 \\
 & Factoid & 0.173 & 0.825 & 0.761 \\
 & List & 0.220 & 0.393 & 0.305 \\
 & MC & 0.782 & 0.916 & 0.866 \\
 & Synthesis & 0.276 & 0.401 & 0.399 \\
\bottomrule
\end{tabular}
\caption{Test-split F1 by question type (3 d.p., $[0,1]$) under zero-shot (ZS), oracle, and RAG ($n=826$).}
\label{tab:type_breakdown}
\end{table}

\section{Human Evaluation}
\label{sec:human_eval}
AfriEconQA includes a stratified human-evaluation subset of 200 examples
sampled from the 4{,}138-instance pool (not from a single split), with
40 examples from each question type (seed 42). This evaluation audits
\textit{benchmark quality}, not model outputs. Independent evaluators inspect
each item against the source PDF and label answer validity as
\texttt{correct}, \texttt{partial}, \texttt{incorrect}, or
\texttt{unanswerable}. Three independent evaluators each reviewed all 200
items, producing 600 labels. Agreement remained stable as annotation expanded.
The first pair agreed on \textbf{89\%} of items, with Gwet's AC1 at
\textbf{0.85}. Across all three annotator pairs, observed agreement was
\textbf{88\%} (530 of 600 pairwise comparisons), with an AC1 of
\textbf{0.88}. All three evaluators assigned the same label to \textbf{83\%}
of items (166/200). We report AC1 because it is less sensitive than
$\kappa$-type statistics to the strong prevalence of \texttt{correct}
labels~\cite{feinstein1990high}.

Across the 600 individual judgments, \textbf{92\%} were \texttt{correct}.
Majority voting resolved 198 of 200 items, and \textbf{95\%} of those
majorities were \texttt{correct} (189/198). The 95\% Wilson confidence
interval is $[92\%,98\%]$.
Majority-correct rates ranged from 92\% to 98\% across question types. This
evaluation audits whether gold answers are supported by the source PDF rather
than estimating human performance on the QA task.

\subsection{Dataset Explorer}
\begin{figure}[ht]
\centering
\includegraphics[width=\columnwidth]{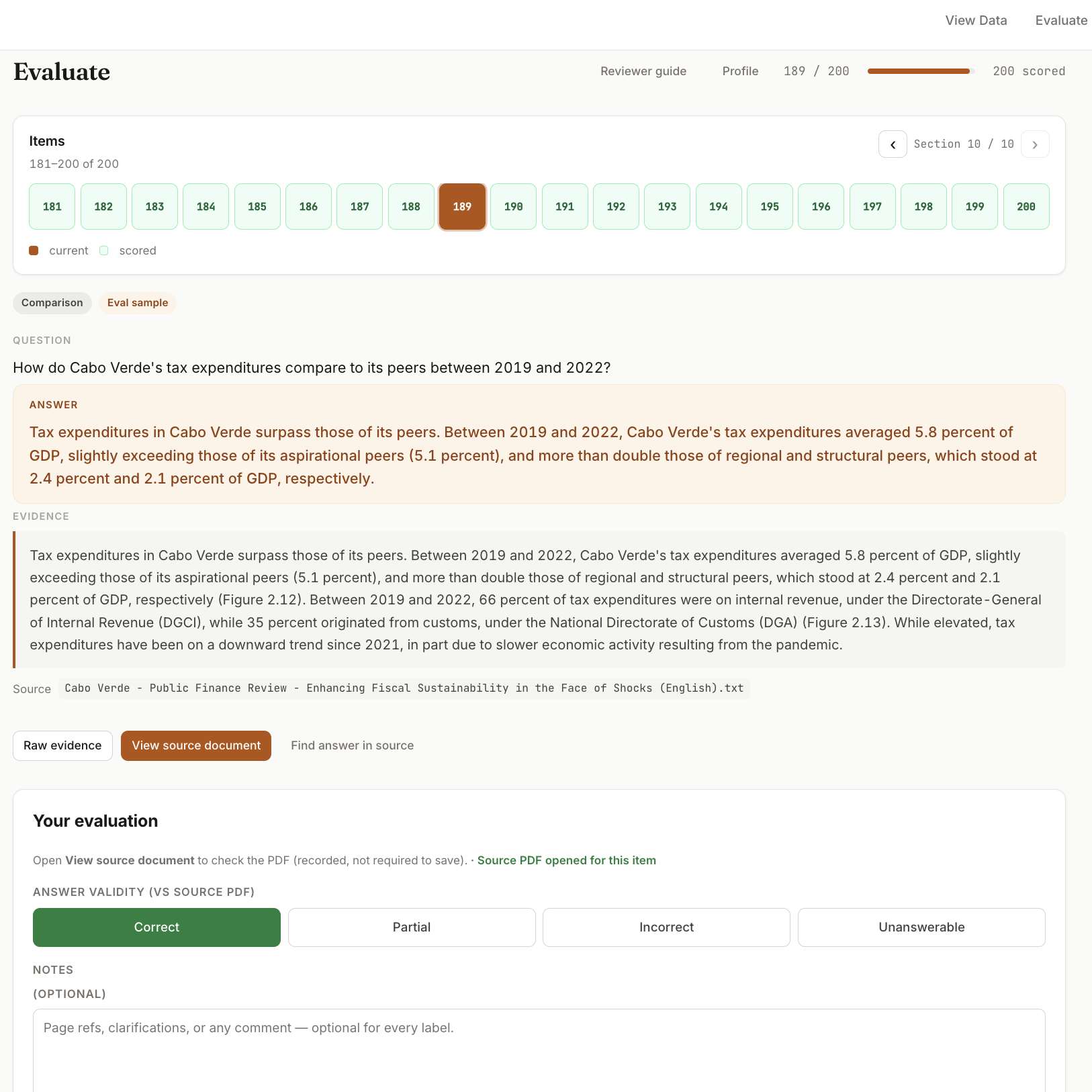}
\caption{AfriEconQA explorer audit view: question, gold answer, evidence, and validity labels.}
\label{fig:dataset_explorer}
\end{figure}

\begin{figure}[ht]
\centering
\includegraphics[width=\columnwidth]{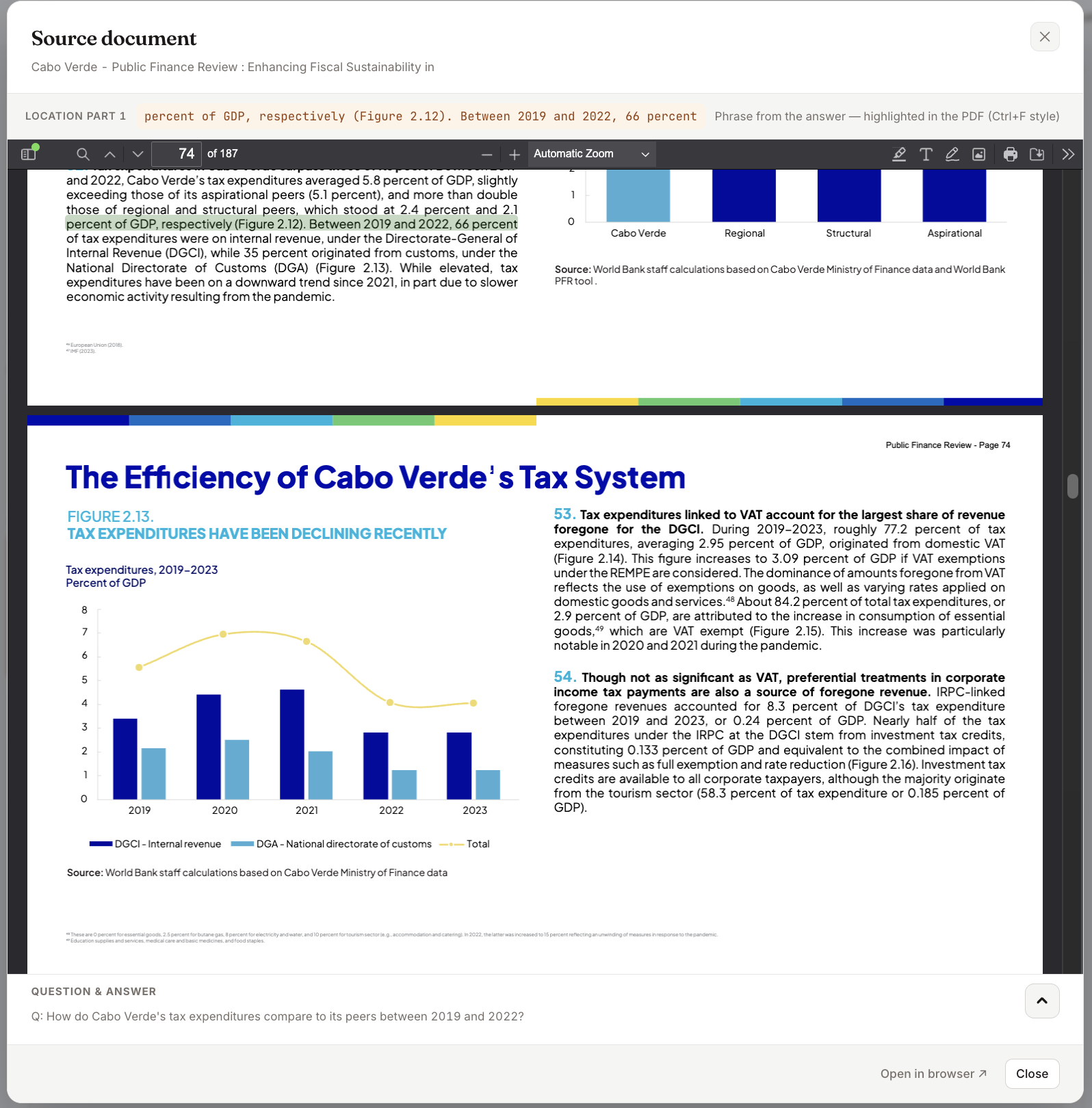}
\caption{Source-PDF view in the explorer, with supporting evidence span highlighted in the World Bank report.}
\label{fig:dataset_pdf}
\end{figure}

To ensure the benchmark is fully auditable and transparent, we provide an
interactive dataset explorer used for sample inspection and human audit
(Figures~\ref{fig:dataset_explorer} and~\ref{fig:dataset_pdf}, Appendix~\ref{app:human_eval}).
The interface exposes the question, gold answer, supporting evidence, source-document
metadata, and the corresponding location in the source PDF where available. It
is the same interface used for the stratified human audit, allowing researchers
to inspect how gold-label quality is checked against the source document. This
serves as a transparency mechanism to verify the benchmark's grounding assumptions
rather than treating the dataset as an opaque file.

\section{Error Analysis}
Two failure modes dominate. First, retrieval often finds the right report but
not the exact claim: document-level HR@5 is high, while chunk-level HR@5
remains near 0.5 (Table~\ref{tab:retriever_results}). World Bank reports
reuse the same indicators across countries, editions, and horizons, so
topically related chunks are frequent near misses. Second, generation remains
hard when the answer must be structured. List questions stay weak under both
oracle and RAG, so missing evidence alone does not explain the errors. Models
also fail at complete set recovery. Together with Table~\ref{tab:type_breakdown},
these patterns show that AfriEconQA stresses evidence localization and
structured answer generation more than topical document retrieval.

\section{Conclusion}
We introduced AfriEconQA, a benchmark of 4,138 evidence-linked questions for
document-grounded QA over World Bank economic reports. The benchmark targets
quantitative, temporal, and structural reasoning over institutional documents
whose claims are difficult to answer from parametric memory alone. Across
four models and three controlled conditions, zero-shot performance
remains low, RAG substantially improves answer quality, and type-level analysis
reveals persistent weaknesses on List, Synthesis, and Comparison questions.

These findings show that reliable economic-report QA requires more than
topical retrieval: systems must localize country- and time-scoped evidence,
preserve numerical values, and transform complex policy prose into structured
answers. AfriEconQA provides an evidence-grounded testbed for these capabilities.

\clearpage
\section*{Limitations}
AfriEconQA is scoped to World Bank economic reports on African economies from
2024--2025, so results should not be read as covering all economic document
QA. Gold answers reflect the cited report edition, while tracking later
revisions is out of scope. Human validation audits gold-label support against
the source PDF on a stratified 200-item subset. It does not measure human
performance on the full open-ended QA task, which would require different
expertise and protocol. The LLM judge is a secondary semantic signal and can be more
lenient than exact match on long answers. Oracle uses a single gold evidence
passage while RAG uses top-five chunks, so oracle is a controlled evidence-use
condition rather than a strict performance ceiling. Stronger specialized
retrievers and longer-context generators may further improve absolute scores.
Our contribution is the benchmark and the controlled evaluation protocol.

\section*{Reproducibility and Open Release}
AfriEconQA is released under a Creative Commons Attribution 4.0
(CC BY 4.0) license. Source World Bank reports remain under World Bank
Open Access / CC BY 4.0 terms. Each instance retains a source URL for
attribution, and the original PDFs are not redistributed. The release includes
the evidence-linked QA file, deterministic type-stratified
train/validation/test splits (seed 42, with primary evaluation on test,
$n=826$), the stratified 200-item human-audit subset, and evaluation scripts
under an open-source license. An interactive dataset explorer is hosted
anonymously at \url{https://afrieconqa.pages.dev/}. Prompts are in Appendix~\ref{app:prompts},
and fixed implementation settings are in Appendix~\ref{app:impl} and in the supplementary Code and Data package.

\clearpage

\bibliography{custom}

\clearpage
\appendix

\section*{Appendix: Table of Contents}
\addcontentsline{toc}{section}{Contents of the Appendix}

\newcommand{\apptoc}[2]{\noindent\textbf{#1} \dotfill \pageref{#2}\\[0.35ex]}
\newcommand{\apptocsub}[2]{\noindent\hspace{1.5em}#1 \dotfill \pageref{#2}\\[0.35ex]}

\apptoc{\ref{app:full_results} Full-Benchmark Results}{app:full_results}
\apptoc{\ref{app:impl} Implementation Details}{app:impl}
\apptoc{\ref{app:prompts} Prompts}{app:prompts}
\apptocsub{\ref{app:prompts_qa} QA Generation Prompts}{app:prompts_qa}
\apptocsub{\ref{app:prompts_eval} Evaluation Prompts}{app:prompts_eval}
\apptocsub{\ref{app:prompts_judge} LLM-as-a-Judge Evaluation Prompt}{app:prompts_judge}
\apptoc{\ref{app:error_analysis} Qualitative Error Analysis and Audited Failure Modes}{app:error_analysis}
\apptocsub{\ref{app:error_modes} Failure Modes in Multi-Item Extraction}{app:error_modes}
\apptocsub{\ref{app:rag_collisions} Indicator and Temporal Collisions in RAG Retrieval}{app:rag_collisions}
\apptocsub{\ref{app:synthesis_omission} Causal Mechanism Omission in Synthesis Reasoning}{app:synthesis_omission}
\apptoc{\ref{app:human_eval} Human Evaluation Protocol and Audit Breakdown}{app:human_eval}
\apptocsub{\ref{app:human_eval_audit} Label Breakdown and Majority-Vote Resolution}{app:human_eval_audit}
\apptocsub{\ref{app:human_eval_explorer} Dataset Explorer and Inspection Interface}{app:human_eval_explorer}

\vspace{0.8em}
\noindent\textbf{AfriEconQA Dataset Explorer:}\\
\url{https://afrieconqa.pages.dev/}
\vspace{1.2em}

\onecolumn
\section{Full-Benchmark Results}
\label{app:full_results}
Primary results in the main paper use the held-out test split ($n=826$).
For completeness, Tables~\ref{tab:main_results_full}--\ref{tab:type_breakdown_full}
report the same metrics on the full benchmark ($n=4{,}138$). Relative model
rankings and the document-vs-chunk localization gap are unchanged.

\begin{table}[ht]
\centering
\small
\begin{tabular}{llcccc}
\toprule
\textbf{Model} & \textbf{Condition} & \textbf{EM} & \textbf{F1} & \textbf{ROUGE-L} & \textbf{LLM-judge} \\
\midrule
\multirow{3}{*}{\textbf{Qwen}} & Zero-shot & 0.162 [0.151, 0.173] & 0.228 [0.218, 0.239] & 0.174 & 0.411 \\
 & Oracle & 0.392 [0.377, 0.408] & 0.494 [0.480, 0.508] & 0.286 & 0.605 \\
 & RAG & 0.426 [0.411, 0.440] & 0.537 [0.524, 0.550] & 0.314 & 0.708 \\
\midrule
\multirow{3}{*}{\textbf{DeepSeek}} & Zero-shot & 0.190 [0.178, 0.201] & 0.259 [0.247, 0.270] & 0.192 & 0.477 \\
 & Oracle & 0.375 [0.360, 0.389] & 0.474 [0.461, 0.488] & 0.279 & 0.572 \\
 & RAG & 0.439 [0.424, 0.453] & 0.551 [0.537, 0.563] & 0.319 & 0.721 \\
\midrule
\multirow{3}{*}{\textbf{Gemma}} & Zero-shot & 0.159 [0.148, 0.170] & 0.224 [0.213, 0.234] & 0.169 & 0.404 \\
 & Oracle & 0.327 [0.312, 0.340] & 0.416 [0.403, 0.430] & 0.252 & 0.517 \\
 & RAG & 0.415 [0.400, 0.429] & 0.523 [0.511, 0.537] & 0.307 & 0.702 \\
\bottomrule
\end{tabular}
\caption{Full-benchmark main results ($n=4{,}138$). All metrics in $[0,1]$. EM and F1 include 95\% bootstrap CIs. LLM-judge is the mean Groq-hosted Llama-3.3-70B LLM-as-a-judge score (secondary).}
\label{tab:main_results_full}
\end{table}

\begin{table}[ht]
\centering
\small
\begin{tabular}{@{}lccc@{}}
\toprule
\textbf{Metric} & \textbf{BM25} & \textbf{BGE-M3} & \textbf{Hybrid} \\
\midrule
HR@1 & 0.271 & 0.307 & 0.325 \\
HR@5 & 0.432 & 0.490 & 0.496 \\
MRR@5 & 0.333 & 0.379 & 0.392 \\
Doc@1 & 0.741 & 0.805 & 0.825 \\
Doc@5 & 0.902 & 0.945 & 0.954 \\
Doc MRR@5 & 0.806 & 0.864 & 0.879 \\
\bottomrule
\end{tabular}
\caption{Full-benchmark retriever-only results ($n=4{,}138$). All metrics in $[0,1]$.}
\label{tab:retriever_results_full}
\end{table}

\begin{table}[ht]
\centering
\footnotesize
\begin{tabular}{@{}llccc@{}}
\toprule
\textbf{Model} & \textbf{Type} & \textbf{ZS} & \textbf{Oracle} & \textbf{RAG} \\
\midrule
\multirow{5}{*}{\textbf{Qwen}} & Comparison & 0.129 & 0.369 & 0.486 \\
 & Factoid & 0.050 & 0.579 & 0.705 \\
 & List & 0.006 & 0.154 & 0.084 \\
 & MC & 0.677 & 0.926 & 0.876 \\
 & Synthesis & 0.276 & 0.332 & 0.441 \\
\midrule
\multirow{5}{*}{\textbf{DeepSeek}} & Comparison & 0.161 & 0.342 & 0.507 \\
 & Factoid & 0.118 & 0.500 & 0.726 \\
 & List & 0.007 & 0.165 & 0.085 \\
 & MC & 0.706 & 0.940 & 0.892 \\
 & Synthesis & 0.288 & 0.324 & 0.445 \\
\midrule
\multirow{5}{*}{\textbf{Gemma}} & Comparison & 0.117 & 0.297 & 0.490 \\
 & Factoid & 0.042 & 0.474 & 0.708 \\
 & List & 0.004 & 0.140 & 0.077 \\
 & MC & 0.679 & 0.793 & 0.815 \\
 & Synthesis & 0.270 & 0.282 & 0.436 \\
\bottomrule
\end{tabular}
\caption{Full-benchmark F1 by question type ($n=4{,}138$, 3 d.p., $[0,1]$).}
\label{tab:type_breakdown_full}
\end{table}

\clearpage
\twocolumn

\section{Implementation Details}
\label{app:impl}
Table~\ref{tab:impl_settings} lists the fixed settings used for the reported
experiments. Each generator is evaluated once per condition on the held-out
test split ($n=826$). Hyperparameters were not tuned on the test split.
The supplementary Code and Data package ships the same values in
\texttt{config.py}.

\begin{table}[t]
\centering
\small
\begin{tabular}{@{}p{0.42\linewidth}p{0.50\linewidth}@{}}
\toprule
\textbf{Setting} & \textbf{Value} \\
\midrule
\multicolumn{2}{@{}l@{}}{\textit{Generators}} \\
Qwen 3.6 35B & \texttt{qwen/qwen3.6-35b-a3b} (LM Studio) \\
Gemma 4 12B IT & \texttt{google/gemma-4-12b} (LM Studio) \\
DeepSeek v4-pro & \texttt{deepseek-v4-pro} (DeepSeek API) \\
GPT-5.5 & \texttt{gpt-5.5} (LiteLLM) \\
Decoding (all generators) & temperature 0.2, top-p 0.95, max 320 tokens \\
Evaluation runs & one pass per generator $\times$ condition \\
\midrule
\multicolumn{2}{@{}l@{}}{\textit{Retrieval (RAG)}} \\
Retriever & BM25 + BGE-M3 (\texttt{BAAI/bge-m3}) \\
Fusion & Reciprocal Rank Fusion, $c{=}60$ \\
Candidate depth / top-$k$ & 60 / 5 \\
Index chunk size & 1{,}000 characters, 200 overlap \\
\midrule
\multicolumn{2}{@{}l@{}}{\textit{Metrics and splits}} \\
LLM judge & Gemini 3.7 Flash, temp.\ 0, max 2048 tokens \\
Bootstrap CIs (EM/F1) & 1{,}000 resamples, seed 42 \\
Train/val/test splits & seed 42 (2{,}896 / 416 / 826) \\
\bottomrule
\end{tabular}
\caption{Fixed implementation settings (also in supplementary \texttt{config.py}).}
\label{tab:impl_settings}
\end{table}

\section{Prompts}
\label{app:prompts}
We list the main prompts used for dataset generation, model evaluation, and
LLM judging. Minor wording may vary by question type. The templates below
are the ones used for the reported experiments.

\subsection{QA Generation Prompts}
\label{app:prompts_qa}
Questions are produced by five type-specific agents
(\texttt{Llama-3.3-70B-Instruct}). Each agent uses a system prompt with a
role line, shared critical rules, and a JSON output schema, plus a user
message that provides the report chunk.

\begin{promptbox}{Agent Role Lines}
\promptlabel{Factoid}
You are a World Bank Data Agent. Extract Factoid QA pairs with precise
numerical or categorical facts. Only generate if the text has strong,
unambiguous statistical claims.

\smallskip
\promptlabel{List}
You are a World Bank Synthesis Agent. Extract List QA pairs identifying
multiple discrete factors, risks, or policy drivers. Only generate if the
text explicitly enumerates multiple items.

\smallskip
\promptlabel{Comparison}
You are a World Bank Analytical Agent. Extract Comparison QA pairs reasoning
across time periods or regions. Only generate if the text explicitly
compares data across years or countries.

\smallskip
\promptlabel{Multiple Choice}
You are a World Bank Assessment Agent. Generate Multiple Choice QA pairs
with 4 options (A/B/C/D). Only generate if the text supports strong
distractor construction.

\smallskip
\promptlabel{Synthesis}
You are a World Bank Macroeconomic Agent. Extract Synthesis QA pairs
covering complex causal chains and economic transmission mechanisms. Only
generate if the text explicitly describes how policy X caused outcome Y.
\end{promptbox}

\begin{promptbox}{Shared Critical Rules (System)}
\begin{enumerate}\setlength\itemsep{2pt}
	\item DO NOT force questions. If the text chunk does not support good
	questions for this task, return \texttt{[]}.
	\item Quality over quantity. Some chunks can produce 0 questions.
	\item Questions must be self-contained and specific (name
	country/region/time period whenever available in the text).
	\item DO NOT use meta-language like ``According to the report'' or ``The
	document states''.
	\item Use semantic understanding: question, answer, and
	\texttt{evidence} may be paraphrased.
	\item If one sentence in the TEXT CHUNK fully supports the answer and
	contains enough context, copy that sentence exactly into
	\texttt{evidence}.
	\item If the answer requires combining multiple sentences, figures,
	table rows, or context from different parts of the chunk, write a
	concise synthesized evidence paragraph in \texttt{evidence}.
	\item Whenever \texttt{evidence} is synthesized, include the original
	verbatim supporting sentences or table rows in \texttt{raw\_evidence}.
	\item \texttt{raw\_evidence} must always contain only verbatim text
	copied from the TEXT CHUNK.
	\item Do not require the answer wording to exactly match the evidence
	wording. Semantically equivalent wording is acceptable if grounded in
	\texttt{raw\_evidence}.
	\item For table-based questions, include row/value text. Do not use
	only the table title.
	\item \texttt{evidence\_rationale} must be exactly one sentence and 40
	words or fewer.
\end{enumerate}
\end{promptbox}

\begin{promptbox}{Standard JSON Output Schema (non-MCQ)}
\begin{lstlisting}[style=promptjson]
Output strictly as a raw JSON array. No markdown, no explanation text before or after.

[
  {
    "question_type": "{q_type}",
    "question": "...",
    "answer": "...",
    "evidence": ["Either one exact supporting sentence OR a synthesized
      evidence paragraph grounded in raw_evidence"],
    "raw_evidence": ["verbatim supporting sentence/table row copied
      from TEXT CHUNK"],
    "evidence_rationale": "Exactly 1 sentence and 40 words or less
      explaining why this evidence supports the answer."
  }
]
\end{lstlisting}
\end{promptbox}

\begin{promptbox}{Multiple Choice Extras}
Appended to the 12 shared rules above for the Multiple Choice agent only,
replacing the standard output schema.
\begin{lstlisting}[style=promptjson]
10. MULTIPLE CHOICE FORMAT: The output JSON for each MC item MUST include an
"options" field as a dictionary with exactly four keys: "A", "B", "C", "D".
Exactly one option is the correct answer. The other three are STRONG
DISTRACTORS -- they must be wrong, but plausible to someone who has not read
the report carefully (e.g., similar numbers, related-but-wrong institutions,
plausible-but-incorrect causal chains). Avoid obviously absurd distractors.

11. Do NOT always place the correct answer in option A. Randomize the
correct option position across questions.

Output format for Multiple Choice:
[
  {
    "question_type": "Multiple Choice",
    "question": "...",
    "answer": "B",
    "options": {"A": "...", "B": "[CORRECT ANSWER TEXT]", "C": "...",
      "D": "..."},
    "evidence": ["direct sentence or concise synthesis..."],
    "raw_evidence": ["exact source sentence/span from chunk..."],
    "evidence_rationale": "One sentence, <= 40 words."
  }
]
\end{lstlisting}
\end{promptbox}

\begin{promptbox}{User Message Template}
REPORT FILENAME: $\langle$filename$\rangle$ (Part $i$/$N$)

TEXT CHUNK:\\
$\langle$chunk text$\rangle$

Generate the $\langle$Type$\rangle$ JSON QA pairs now if applicable,
otherwise return \texttt{[]}.
\end{promptbox}

\subsection{Evaluation Prompts}
\label{app:prompts_eval}
\begin{promptbox}{Zero-Shot Evaluation Prompt}
Answer the following question about African macroeconomic policy
accurately. Keep your answer concise and direct.

\smallskip
Question: $\langle$question$\rangle$\\
Answer:

\smallskip
\textit{(MCQ)} Answer with the correct option letter (A, B, C, or D)
followed by a short answer phrase. Do not include extra commentary.\\
\textit{(other)} Answer in a single concise sentence with no extra
commentary.
\end{promptbox}

\begin{promptbox}{Oracle Evaluation Prompt}
Using the provided context, answer the following question about African
macroeconomic policy accurately. Keep your answer concise and direct. If the
context does not contain the information needed to answer the question,
output ``Information not provided''.

\smallskip
Context:\\
$\langle$gold evidence$\rangle$

\smallskip
Question: $\langle$question$\rangle$\\
Answer: \textit{(same answer-format instruction as zero-shot)}
\end{promptbox}

\begin{promptbox}{RAG Evaluation Prompt}
Using the provided retrieved documents, answer the following question about
African macroeconomic policy accurately. Keep your answer concise and
direct. If the documents do not contain the information needed to answer
the question, output ``Information not provided''.

\smallskip
Documents:\\
$\langle$top-5 retrieved chunks$\rangle$

\smallskip
Question: $\langle$question$\rangle$\\
Answer: \textit{(same answer-format instruction as zero-shot)}
\end{promptbox}

\subsection{LLM-as-a-Judge Evaluation Prompt}
\label{app:prompts_judge}
The judge model is \texttt{Llama-3.3-70B-Instruct} on Groq, called once per
prediction with a fixed system prompt and a per-item user prompt built from
the template below.

\begin{promptbox}{Judge System Prompt}
You are evaluating a generated answer against a reference answer for a
factual QA benchmark. Score factual agreement with the reference only. Do
not reward an answer for discussing the correct topic, sounding fluent, or
being verbose. Reward it only for stating the same facts. A claim that contradicts
the reference is worse than one it simply omits. Treat the reference as
ground truth: do not use outside knowledge, and do not decide the reference
itself is wrong. Output JSON only.
\end{promptbox}

\begin{promptbox}{Per-Type Verification Instruction}
One line is substituted for \texttt{\{task\_prompt\}} in the user prompt
below, selected by question type.
\begin{itemize}\setlength\itemsep{2pt}
	\item \textbf{factoid:} Verify that the answer gives the correct entity
	or value, including any essential unit and time qualifier.
	\item \textbf{multiple\_choice:} Verify that the answer selects the
	correct option. Compare the generated answer to the gold option TEXT
	exactly. If it states a value or phrase that matches a different
	option, score 0 even if the topic is right.
	\item \textbf{list:} Verify that the answer includes the required
	items. Give partial credit when only some required items are correct.
	\item \textbf{synthesis:} Verify the main conclusion and supporting
	relationship or mechanism. Topic overlap alone is not enough.
	\item \textbf{comparison:} Verify the compared entities, direction of
	the comparison, and any stated values or magnitudes.
	\item \textbf{default:} Verify that the generated answer correctly and
	completely answers the question.
\end{itemize}
\end{promptbox}

\begin{promptbox}{Judge User Prompt Template}
\begin{lstlisting}[style=promptjson]
{task_prompt}

Question: {question}
Reference Answer: {truth}
Generated Answer: {prediction}
{options_block}

Score using this scale:
10 = fully correct and complete, with no missing or incorrect facts
8-9 = correct, with only a minor imprecision in phrasing and all facts correct
6-7 = mostly correct, missing one relevant fact but stating nothing that
      contradicts the reference
4-5 = missing more than one relevant fact, but nothing stated contradicts
      the reference
2-3 = contains a factual contradiction, or is on-topic with no correct
      facts
0-1 = incorrect, contradictory throughout, irrelevant, empty, or a
      refusal

A contradicted fact always scores in the 0-3 range, even if other facts in
the same answer are correct -- never average a contradiction against
surrounding correct facts.

List only facts explicitly stated in the reference or generated answer for
matched/missing/incorrect below -- do not infer implications, and do not
add facts that appear in neither answer.

Return exactly {"score": <integer 0-10>, "matched": ["..."], "missing":
["..."], "incorrect": ["..."], "reasoning": "<one sentence>"}.
\end{lstlisting}
For Multiple Choice items, \texttt{\{options\_block\}} additionally lists
the four options and the gold correct option letter. Reported LLM-judge scores
normalize the returned integer to $[0,1]$ by dividing by 10.
\end{promptbox}

\section{Qualitative Error Analysis and Audited Failure Modes}
\label{app:error_analysis}

\begin{table}[ht]
\centering
\small
\resizebox{\columnwidth}{!}{%
\begin{tabular}{@{}lp{0.82\columnwidth}@{}}
\toprule
\textbf{Case \& Type} & \textbf{Audit Comparison (Gold vs.\ Predictions)} \\
\midrule
\textbf{ID 193} & \textbf{Q:} What is the inflation rate of Somalia in 2024? \\
Factoid & \textbf{Gold:} 5.6 percent \\
        & \textbf{Oracle:} Somalia's inflation rate in 2024 was 5.6 percent. \\
        & \textbf{RAG (Near-Miss):} Somalia's inflation rate averaged 5.5 percent in 2024. \\
\midrule
\textbf{ID 549} & \textbf{Q:} Projected poverty rate in Ghana in 2025 (LMIC line)? \\
Factoid & \textbf{Gold:} 53.3 percent \\
        & \textbf{Oracle:} The projected poverty rate was 53.3 percent. \\
        & \textbf{RAG (Near-Miss):} The projected poverty rate was 53.8 percent. \\
\midrule
\textbf{ID 462} & \textbf{Q:} Malawi 2050 GDP loss under BAU climate scenario? \\
Factoid & \textbf{Gold:} 15.8 percent \\
        & \textbf{Oracle:} Estimated annual GDP loss is 15.8 percent below baseline. \\
        & \textbf{RAG (Near-Miss):} Annual GDP loss under BAU is estimated at about 15.7 percent. \\
\midrule
\textbf{ID 63}  & \textbf{Q:} What is the 'oil for roads model' in South Sudan? \\
Synthesis & \textbf{Gold:} Financing model allocating oil revenue for road development. \\
        & \textbf{Oracle:} Road initiative starting with Juba-Bor and Juba-Terekeka roads. \\
        & \textbf{RAG (Near-Miss):} Transport initiative initiated in 2019 for regional connectivity. \\
\midrule
\textbf{ID 79}  & \textbf{Q:} Components of Guinea-Bissau Human Capital Project? \\
List    & \textbf{Gold:} 5 components (women/children, services, HR, M\&E, contingency). \\
        & \textbf{Oracle:} Lists 3 components and concludes with generic summary. \\
        & \textbf{RAG (Near-Miss):} Lists 2 components and omits contingency response. \\
\bottomrule
\end{tabular}%
}
\caption{Audited qualitative failure cases from the held-out test split ($n=826$) demonstrating near-miss RAG retrieval collisions and Oracle synthesis omissions.}
\label{tab:audited_cases}
\end{table}

To understand why performance varies sharply across question types, we conducted
a qualitative error analysis across model predictions on the held-out test split ($n=826$).
Our analysis isolates three distinct empirical phenomena that explain the observed
performance ceilings without reflecting annotation artifacts.

\subsection{Failure Modes in Multi-Item Extraction}
\label{app:error_modes}

\paragraph{1. Premature Stopping and Truncation in Multi-Item Sets:}
The most frequent error mode in List questions is incomplete set enumeration.
When questions inquire about multi-component policy packages, sector constraints,
or strategic pillars (e.g., the five components of the Guinea-Bissau Human Capital
Project, ID 79, or the key determinants of poverty reduction in Liberia, ID 3822),
models routinely extract the first two or three items and terminate generation,
often appending concluding phrases such as ``among others'' or ``and related reforms''.
While the emitted items are factually correct, the omitted items cause substantial
penalties in set recall. This pattern persists across all model families, indicating
an inherent generator tendency toward conciseness over exhaustive set enumeration.

\paragraph{2. Granularity and Institutional Title Divergence:}
Macroeconomic reports frequently refer to specialized governmental entities,
legal directives, or project indicators by formal statutory titles (e.g.,
``National Social Protection Commission'' or ``Job-rich growth and data systems'').
Models often rephrase these formal titles into broader conceptual summaries
(e.g., ``welfare initiatives'' or ``economic expansion''). While semantic intent
is preserved, token-level overlap and exact string matching are substantially
penalized. This divergence is especially pronounced under RAG, where surrounding
commentary in retrieved chunks tempts models into narrative summarization rather
than crisp entity extraction.

\paragraph{3. Context Completeness and Model Refusal:}
A critical factor governing extraction performance is context sufficiency.
When models are explicitly instructed to avoid hallucination and output
``Information not provided'' if evidence is absent, fragmented or under-specified
evidence spans trigger conservative non-response refusals rather than partial guesses.
Ensuring that supporting evidence captures the complete verbatim paragraph
allows models to ground multi-item queries directly in source context,
minimizing spurious refusals.

\subsection{Indicator and Temporal Collisions in RAG Retrieval}
\label{app:rag_collisions}
In macroeconomic reporting, identical indicators recur across adjacent fiscal years,
forecast revisions, or counterfactual policy projections within the same report.
Dense embedding models and sparse lexical matchers readily identify the correct document
(Doc@5 reaches 0.954 on the full benchmark), but they struggle to differentiate
neighboring paragraphs whose semantic content differs solely by a temporal anchor or scenario label.

As shown in Table~\ref{tab:audited_cases}, this produces systematic near-miss retrieval errors:
\begin{itemize}
    \item \textbf{Adjacent Year Collisions (ID 193):} When asked \emph{``What is the inflation rate of Somalia in 2024?''}, the gold answer is \textbf{5.6\%} (the year-end December figure). However, top-5 RAG retrieved an adjacent paragraph discussing historical baseline averages (\textbf{5.5\%}). Because the two passages share near-identical phrasing except for the temporal context, the generator bound to the wrong figure and output \emph{``Somalia's inflation rate averaged 5.5 percent in 2024''}.
    \item \textbf{Multi-Horizon Projection Collisions (ID 549):} When asked for the \emph{``Projected poverty rate in Ghana in 2025 (LMIC line)''}, the gold projection is \textbf{53.3\%}. RAG retrieved an adjacent intermediate planning target from the same report (\textbf{53.8\%}). The model generated \emph{``The projected poverty rate was 53.8 percent''}, missing the gold target by just 0.5\% due to temporal misalignment in retrieved chunks.
    \item \textbf{Cross-Scenario Policy Collisions (ID 462):} When asked for \emph{``Malawi 2050 GDP loss under BAU climate scenario?''}, the gold figure is \textbf{15.8\%} under the baseline business-as-usual trajectory. The retriever pulled an adjacent table row from the same report describing the alternative \emph{Wet} scenario (\textbf{15.7\%}), causing the model to predict \emph{``15.7 percent''} instead of 15.8\%.
\end{itemize}

\subsection{Causal Mechanism Omission in Synthesis Reasoning}
\label{app:synthesis_omission}
On Synthesis and Comparison questions, models exhibit surface entity fixation even when provided
with the complete gold context in the Oracle condition.
For example, in the South Sudan transport diagnostic (ID 63), the query asks to synthesize the
`oil for roads model'. The gold answer defines the institutional financing arrangement wherein
oil revenues are allocated to fund road infrastructure. Under Oracle context, frontier generators
frequently fixate on the specific road segments mentioned in the passage (such as the 198 km Juba-Bor road)
while omitting the substantive macroeconomic financing mechanism entirely.
Similarly, in ID 62, models latch onto general agricultural diversification background sentences
while dropping the core social transformation and marginalization clauses demanded by the inquiry.
These qualitative instances demonstrate that benchmark difficulty arises from genuine macroeconomic
reasoning demands rather than noisy gold annotations.

\section{Human Evaluation Protocol and Audit Breakdown}
\label{app:human_eval}
To provide full transparency on the gold-label quality audit (Section~\ref{sec:human_eval}),
Table~\ref{tab:human_eval_breakdown} reports the complete distribution across all 600
individual judgments and majority-vote classifications on the 200 triple-annotated items.

\begin{table}[ht]
\centering
\small
\resizebox{\columnwidth}{!}{%
\begin{tabular}{lcccc}
\toprule
\textbf{Label} & \textbf{Majority ($n=200$)} & \textbf{Resolved (\%)} & \textbf{Labels ($N=600$)} & \textbf{Label (\%)} \\
\midrule
Correct & 189 & 95.5\% & 552 & 92.0\% \\
Partial & 5 & 2.5\% & 33 & 5.5\% \\
Incorrect & 3 & 1.5\% & 10 & 1.7\% \\
Unanswerable & 1 & 0.5\% & 5 & 0.8\% \\
Ties (no majority) & 2 & 1.0\% & -- & -- \\
\midrule
\textbf{Total} & \textbf{200} & \textbf{100.0\%} & \textbf{600} & \textbf{100.0\%} \\
\bottomrule
\end{tabular}%
}
\caption{Distribution of human evaluation judgments across all 600 labels and majority-vote outcomes ($n=200$).}
\label{tab:human_eval_breakdown}
\end{table}

\subsection{Label Breakdown and Majority-Vote Resolution}
\label{app:human_eval_audit}

\paragraph{Data Quality Score Calculation:}
To calculate the 95.5\% data quality score ($189/198$, Wilson 95\% CI $[92\%, 98\%]$), only items judged fully correct are counted as successes. Any item judged as \texttt{partial} (5), \texttt{incorrect} (3), or \texttt{unanswerable} (1) is counted strictly as an error, with no partial credit given.

\paragraph{Dataset Disposition and Audit Role:}
The stratified 200-item human evaluation functions as an independent quality audit to measure
the empirical accuracy of the generation pipeline across the benchmark, rather than an ad-hoc
filtering mechanism. None of the audited items were discarded or altered post-hoc, ensuring
that reported statistics truthfully reflect the underlying 4{,}138-instance distribution.
Across question types, majority-correct rates remain consistently high: List (39/40, 97.5\%),
Synthesis (38/39, 97.4\%), Multiple Choice (38/40, 95.0\%), Comparison (37/39, 94.9\%),
and Factoid (37/40, 92.5\%).

\subsection{Dataset Explorer and Inspection Interface}
\label{app:human_eval_explorer}
An anonymous interactive deployment of the dataset explorer is accessible
at \url{https://afrieconqa.pages.dev/} for inspecting sampled question-answer
pairs alongside original World Bank source reports.

\end{document}